\documentclass{article}

\usepackage{PRIMEarxiv}

\usepackage[style=acmnumeric]{biblatex} 
\usepackage[utf8]{inputenc} 
\usepackage[T1]{fontenc}    
\usepackage{hyperref}       
\usepackage{url}            
\usepackage{booktabs}       
\usepackage{amsfonts}       
\usepackage{nicefrac}       
\usepackage{microtype}      
\usepackage{lipsum}
\usepackage{fancyhdr}       
\usepackage{graphicx}       

\title{A template for Arxiv Style
\thanks{\textit{\underline{Citation}}: 
\textbf{Authors. Title. Pages.... DOI:000000/11111.}} 
}

\author{
  Yonadav Shavit \\
  Harvard University \\
  \texttt{yonadav@g.harvard.edu}\\
}

\usepackage{xcolor}
\usepackage{siunitx}
\usepackage{amsmath}
\usepackage{amsthm}

\usepackage{array}

\newtheorem{theorem}{Theorem} 
\newtheorem{lemma}{Lemma}
\newtheorem{definition}{Definition}

\definecolor{DarkGreen}{rgb}{0.1,0.5,0.1}
\newcommand{\todo}[1]{\textcolor{DarkGreen}{[TODO: #1]}}

\newcommand{\future}[1]{}

\newcommand{\ot}[1]{\textcolor{purple}{[Outline: ]}}

\newcommand{\monitoringperiod}{T_m}
\newcommand{\trainingperiod}{T}
\newcommand{\totalchips}{C}

\newcommand{\rulecomputeflops}{H}

\newcommand{\hashfunction}{\mathcal{H}}

\title{What does it take to catch a Chinchilla?\\Verifying Rules on Large-Scale Neural Network Training via Compute Monitoring}

\begin{document}
\maketitle
\begin{abstract}
As advanced machine learning systems' capabilities begin to play a significant role in geopolitics and societal order, it may become imperative that (1) governments be able to enforce rules on the development of advanced ML systems within their borders, and (2) countries be able to verify each other's compliance with potential future international agreements on advanced ML development.
This work analyzes one mechanism to achieve this, by monitoring the computing hardware used for large-scale NN training.
The framework's primary goal is to provide governments high confidence that no actor uses large quantities of specialized ML chips to execute a training run in violation of agreed rules.
At the same time, the system does not curtail the use of consumer computing devices, and maintains the privacy and confidentiality of ML practitioners' models, data, and hyperparameters.
The system consists of interventions at three stages:
(1) using on-chip firmware to occasionally save snapshots of the the neural network weights stored in device memory, in a form that an inspector could later retrieve; (2) saving sufficient information about each training run to prove to inspectors the details of the training run that had resulted in the snapshotted weights; and (3) monitoring the chip supply chain to ensure that no actor can avoid discovery by amassing a large quantity of un-tracked chips.
The proposed design decomposes the ML training rule verification problem into a series of narrow technical challenges, including a new variant of the Proof-of-Learning problem [Jia et al. '21].
\end{abstract}

\section{Introduction}\label{s.intro}
Many of the remarkable advances of the past 5 years in deep learning have been driven by a continuous increase in the quantity of \emph{training compute} used to develop cutting-edge models \cite{kaplan2020scaling, hoffman2022training, sevilla2022compute}.
Such large-scale training has been made possible through the concurrent use of hundreds or thousands of specialized accelerators with high inter-chip communication bandwidth (such as Google TPUs, NVIDIA A100 and H100 GPUs, or AMD MI250 GPUs), employed for a span of weeks or months to compute thousands or millions of gradient updates.
We refer to these specialized accelerators as \emph{ML chips}, which we distinguish from consumer-oriented GPUs with lower interconnect bandwidth (e.g., the NVIDIA RTX 4090, used in gaming computers).

This compute scaling trend has yielded models with ever more useful capabilities.
However, these advanced capabilities also bring with them greater dangers from misuse \cite{brundage2018malicious}.
For instance, it is increasingly plausible that criminals may soon be able to leverage heavily-trained code-generation-and-execution models to autonomously identify and exploit cyber-vulnerabilities, enabling ransomware attacks on an unprecedented scale.
\footnote{The 2017 NotPetya worm was estimated to have caused \$10B in damages, and was both far less capable and far more straightforward to disable \cite{greenberg_2018}.}
Even absent malicious intent, rival companies or countries trapped in an AI ``race dynamic'' may face substantial pressure to cut corners on testing and risk-mitigation, in order to deploy high-capability ML systems in the hopes of outmaneuvering their competitors economically or militarily.
The edge-case behaviors of deep learning models are notoriously difficult to debug \cite{hendrycks2021unsolved}, and without thorough testing and mitigation, such bugs in increasingly capable systems may have increasingly severe consequences.
Even when rival parties would all prefer to individually do more testing and risk-mitigation, or even forgo developing particularly dangerous types of ML models entirely \cite{urbina2022dual}, they may have no way to verify whether their competitors are matching their levels of caution.

In the event that such risks do emerge, governments may wish to enforce limits on the large-scale development of ML models.
While law-abiding companies will comply, criminal actors, negligent companies, and rival governments may not, especially if they believe their rule-violations will go unnoticed.
It would therefore be useful for governments to have methods for reliably \emph{verifying} that large-scale ML training runs comply with agreed rules.

These training runs' current need for large quantities of specialized chips leaves a large physical and logistical footprint, meaning that such activities are generally undertaken by sizable organizations (e.g., corporate or governmental data-center operators) well-equipped to comply with potential regulations.
Yet even if the relevant facilities are known, there is no easily-observable difference between training a model for social benefit, and training a model for criminal misuse --- they require the same hardware, and at most differ in the code and data they use.
Given the substantial promise of deep learning technologies to benefit society, it would be unfortunate if governments, in a reasonable attempt to curtail harmful use-cases but unable to distinguish the development of harmful ML models, ended up repressing the development of beneficial applications of ML as well.
Such dynamics are already appearing: the US Department of Commerce's rationale for its October 2022 export controls denying the sale of high-performance chips to the People's Republic of China, while not specific to ML, was based in part on concern that those chips might be used to develop weapons against the United States or commit human rights abuses \cite{bis2022controls}.
If the US and Chinese governments could reach an agreement on a set of permissible beneficial use-cases for export-controlled chips, and had a way to verify Chinese companies' compliance with that agreement, it may be possible to prevent or reverse future restrictions.

Such a system of verification-based checks and balances, distinguishing between ``safe'' and ``dangerous'' ML model training, might seem infeasible.
Yet a similar system has been created before.
At the dawn of the nuclear age, nations faced an analogous problem:
reactor-grade uranium (used for energy) and weapons-grade uranium (used to build nuclear bombs) could be produced using the same types of centrifuges, just run for longer and in a different configuration.
In response, in 1970 the nations of the world adopted the Treaty on the Non-Proliferation of Nuclear Weapons (NPT) and empowered the International Atomic Energy Agency (IAEA) to verify countries' commitments to limiting the spread of nuclear weapons, while still harnessing the benefits of nuclear power. 
This verification framework has helped the world avoid nuclear conflict for over 50 years, and helped limit nuclear weapons proliferation to just 9 countries while spreading the benefits of safe nuclear power to 33 \cite{wikipedia_2023}.
If future progress in machine learning creates the domestic or international political will for enacting rules on large-scale ML development, it is important that the ML community is ready with technical means for verifying such rules.

\subsection{Contributions}

In this paper, we propose a monitoring framework for enforcing rules on the \emph{training} of ML\footnote{Throughout the text, we use ``ML'' to refer to deep-learning-based machine learning, which has been responsible for much of the progress of recent years.} models using large quantities of specialized ML chips.
Its goal is to enable governments to verify that companies and other governments have complied with agreed guardrails on the development of ML models that would otherwise pose a danger to society or to international stability.
The objective of this work is to lay out a possible system design, analyze its technical and logistical feasibility, and highlight important unsolved challenges that must be addressed to make it work.

The proposed solution has three parts:
\begin{enumerate}
    \item To prove compliance, an ML chip owner employs firmware that logs limited information about that chip's activity, with their employment of that firmware attested via hardware features.
    We propose an activity logging strategy that is both lightweight, and maintains the confidentiality of the chip-owner's trade secrets and private data, based on the NN weights present in the device's high-bandwidth memory.
    \item By inspecting and analyzing the logs of a sufficient subset of the chips, inspectors can provably determine whether the chip-owner executed a rules-violating training run in the past few months, with high probability.
    \item Compute-producing countries leverage supply-chain monitoring to ensure that each chip is accounted for, so that actors can't secretly acquire more ML chips and then underclaim their total to hide from inspectors.
\end{enumerate}
The system is compatible with many different rules on training runs (see Section \ref{s.rules}), including those based on the total chip-hours used to train a model, the type of data and algorithms used, and whether the produced model exceeds a performance threshold on selected benchmarks.
To serve as a foundation for meaningful international coordination, the framework aspires to reliably detect violations of ML training rules \emph{even in the face of nation-state hackers attempting to circumvent it}.
At the same time, the system does not force ML developers to disclose their confidential training data or models.
Also, as its focus is restricted to specialized data-center chips, the system does not affect individuals' use of their personal computing devices.

Section \ref{s.problemform} introduces the problem of verifying rules on large-scale ML training. Section \ref{s.overview} provides an overview of the solution, and describes the occasional inspections needed to validate compliance.
Sections \ref{s.onchip}, \ref{s.datacenter},  and \ref{s.supplychain} discuss the interventions at the chip-level, data-center-level, and supply-chain respectively.
Section \ref{s.discussion} concludes with a discussion of the proposal's benefits for different stakeholders, and lays out near-term next steps.

\subsection{Limitations}
\label{s.limits}
The proposed system's usefulness depends on the continued importance of large-scale training to produce the most advanced (and thus most dangerous) ML models, a topic of uncertainty and ongoing disagreement within the ML community.
The framework's focus is also restricted only to training runs executed on specialized data-center accelerators, which are today effectively necessary to complete the largest-scale training runs without a large efficiency penalty.
In Appendix \ref{app.howmuch}, we discuss whether these two trends are likely to continue.
Additionally, hundreds of thousands of ML chips have already been sold, many of which do not have the hardware security features required by the framework, and may not be retrofittable nor even locatable by governments.
These older chips' importance may gradually decrease with Moore's Law.
But combined with the possibility of less-efficient training using non-specialized chips, these unmonitored compute sources present an implicit lower bound on the minimum training run size that can be verifiably detected by the proposed system.
Still, it may be the case that frontier training runs, which result in models with new emergent capabilities to which society most needs time to adapt, are more likely to require large quantities of monitorable compute.

More generally, the framework does not apply to small-scale ML training, which can often be done with small quantities of consumer GPUs.
We acknowledge that the training of smaller models (or fine-tuning of existing large models) can be used to cause substantial societal harm (e.g., computer vision models for autonomous terrorism drones \cite{pledger2021role}).
Separately, if a model is produced by a large-scale training run in violation of a future law or agreement, that model's weights may from then on be copied undetectably, and it can be deployed using consumer GPUs \cite{sheng2023high} (as ML inference requires far lower inter-chip communication bandwidth).
Preventing the proliferation of dangerous trained models is itself a major challenge, and beyond the scope of this work.
More broadly, society is likely to need laws and regulations to limit the harms from bad actors' misusing such ML models.
However, exhaustively \emph{enforcing} such rules at the hardware-level would require surveilling and policing individual citizens' use of their personal computers, which would be highly unacceptable on ethical grounds.
This work instead focuses attention upstream, regulating whether and how the most dangerous models \emph{are created in the first place}.

Lastly, rather than proposing a comprehensive shovel-ready solution, this work provides a high-level solution design.
Its contribution is in isolating a set of open problems whose solution would be sufficient to enable a system that achieves the policy goal.
If these problems prove unsolvable, the system's design will need to be modified, or its guarantees scaled back.
We hope that by providing a specific proposal to which the community can respond, we will initiate a cycle of feedback, iteration, and counter-proposals that eventually culminates in an efficient and effective method for verifying compliance with large-scale ML training rules.

\subsection{Related Work}
This paper joins an existing literature examining the role that compute may play in the governance of AI.
Early work by Hwang \cite{hwang2018computational} highlighted the potential of computing power to shape the social impact of ML.
Concurrent work by Sastry et al. \cite{sastry2023computing} identifies attributes of compute that make it a uniquely useful lever for governance, and provides an overview of policy options.
Closely-related work by Baker \cite{baker2023nuclear} draws lessons from nuclear arms control for the compute-based verification of international agreements on large-scale ML.

Rather than focusing on specific policies, the work proposes a technical platform for verifying many possible regulations and agreements on ML development.
Already, the EU AI Act has proposed establishing risk-based regulations on AI products \cite{veale2021demystifying}, while US senators have proposed an ``Algorithmic Accountability Act'' to oversee algorithms used in critical decisions \cite{chu_2022}, and the Cyberspace Administration of China (CAC) has established an ``algorithm registry'' for overseeing recommender systems \cite{cac_2022}.
Internationally, many previous works have discussed the general feasibility and desirability of AI arms control \cite{reinhold2022arms, docherty_2020, mittelsteadt2021ai}, with \cite{scharre2022artificial} highlighting the importance of verification measures to the success of potential AI arms control regimes.
Past work has also explored the benefits of international coordination on non-military AI regulation \cite{erdelyi2018regulating}.

The proposed solution involves proving that a rule-violating ML training run was \emph{not} done, in part by proving which other training runs \emph{were} done.
The analysis of the latter problem is heavily inspired by the literature on Proof-of-Learning \cite{jia2021proof, fang2022fundamental} (discussed further in Section \ref{s.datacenter}).
Other works has have used tools from cryptography to train NN models securely across multiple parties \cite{wagh2018securenn}, and to securely prove the correctness of NN inference \cite{lee2020vcnn}.
However, these approaches suffer large efficiency penalties and cannot yet be scaled to cutting-edge model training, rendering them nonviable as a method for verifying rules on large-scale training runs.

\section{The Problem: Detecting Violations of Large-Scale ML Training Rules}\label{s.problemform}
We focus on the setting in which one party (the ``Verifier'') seeks to verify that a given set of ML training rules is being followed, and another party (the ``Prover'') is developing the ML system and wants to prove to the Verifier that it is complying with those rules.
The Verifier can request that the Prover take actions, such as disclosing information on training runs, in order to help the Verifier determine the Prover's compliance.
The Prover is a ``covert adversary'' \cite{aumann2007security} -- they may benefit from \emph{violating} the ML training rule, but will only seek to violate the rule \emph{if they can still appear compliant} to the Verifier.
There are two real-world Prover-Verifier relationships we are particularly interested in:
\begin{itemize}
\item \emph{Domestic Oversight}:
    Governments have a clear interest that the ML systems developed by companies operating within their borders comply with certain rules.
    Regulators can level both civil and criminal penalties on organizations caught violating rules, and often require organizations to maintain records that prove regulatory compliance (e.g., financial transaction record-keeping requirements).
    \future{
    Unfortunately, the benefit of domestically regulating training may be limited, as dangerous ML models can be developed and deployed in one country and yet be transported via the internet and deployed within the borders of another.
    Thus, effectively preventing certain ML use-cases in any one country may require international cooperation (similar to the global nature of policing child sexual abuse material) \future{cite \url{https://onlinelibrary.wiley.com/doi/abs/10.1111/1758-5899.12418}}}
\item \emph{International Oversight}:
The most significant types of ML training rules may be those enforced internationally (on companies and governments in multiple countries), and verified by other governments or international bodies.
    These include enforcing globally-beneficial rules (e.g., combatting disinformation), and verifying arms control agreements (e.g., limiting the development of autonomous code-generating cyberweapons).
    There is precedent for countries abiding by international agreements with strict monitoring regimes when they stand to benefit, such as Russia's historically allowing random U.S. inspections of its missiles as a part of the START treaties, in exchange for certainty that the U.S. was abiding by the same missile limits \cite{schenck2011start}. \looseness=-1
\end{itemize}

Thus, the problem we address is: what minimal set of verifiable actions can the Verifier require the Prover to take that would enable the Verifier to detect, with high probability, whether the Prover violated any training rules?

\subsection{What types of rules can we enforce by monitoring ML training?}\label{s.rules}
It is important that standards and agreements on ML training focus on preventing concrete harm, and otherwise leave society free to realize the broad benefits of highly-capable ML systems.
Indeed, there are many types of ML models that should not only be legal to train, but that should open-sourced so that all of society can benefit from them \cite{solaiman2023gradient}.
The proposed framework focuses only on enforcing rules on the training of those more dangerous models whose creation and distribution would substantially harm society or international security.
Indeed, as mentioned in Section \ref{s.limits}, this framework \emph{could not} prevent smaller-scale training of ML models, and thus limits the risk of overreach by authoritarian Verifiers.
Below are some informative properties that a Verifier could determine by monitoring the training process of an ML model:
\begin{itemize}
    \item \emph{Total training compute}, which has proven to be an indicator for ML models' capabilities \cite{kaplan2020scaling, sutton2019bitter}.
    \item \emph{Properties of the training data}, such as whether a language model's text dataset contains code for cybersecurity exploits.
    \item \emph{Properties of the hyperparameters}, such as the fraction of steps trained via reinforcement learning.
    \item \emph{The resulting model's performance on benchmarks designed to elicit its capabilities}, including whether the model's capabilities exceed agreed-on thresholds, and including interactive benchmarks (e.g. finetuning the model on a particular task).
    \item Combinations of the above --- for example, ``if a model was trained on RL-for-code-generation for greater than $X$ FLOPs, then it should not be trained beyond $Y$ performance on $Z$ benchmarks.''
\end{itemize}
Ultimately, these rule thresholds should be selected based on the model capabilities that would result.
Current ``scaling law'' extrapolations are not yet able to reliably predict ML models' downstream capabilities \cite{ganguli2022predictability}, so finding principled methods for deciding on rule-thresholds that achieve desired policy outcomes is an important area for future work.

If a Verifier can reliably detect the aforementioned training run properties, that would allow them to mandate several types of rules, such as:
\begin{itemize}
\item 
\emph{Reporting requirements} on large training runs, to make domestic regulators aware of new capabilities or as a confidence-building measure between companies/competitors \cite{horowitz2021ai}.
\item 
\emph{Bans or approval-requirements} for training runs considered overly likely to result in models that would threaten society or international stability. Approval could be conditioned on meeting additional requirements (e.g., willingness to comply with downstream regulations on model use, increased security to prevent model-theft, greater access for auditors).
\item 
\emph{Requiring that any trained model be modified to include post-hoc safety mitigations} if the unmodified model could be expected to pose a severe accident risk absent those mitigations.
Such safety assessments and mitigations (such as ``Helpful and Harmless'' finetuning \cite{bai2022training}) may involve a prohibitive upfront cost that companies/governments would otherwise avoid.
However, once they have been forced to make the investment and built a less accident-prone model, they may then prefer to use the safer version.
Such rules allow all parties to coordinate spending more resources on safe and responsible innovation, without fearing that their competitors may secretly undercut them by rushing ahead without addressing negative externalities.
\end{itemize}

\subsection{Other Practical Requirements}
There are several other considerations for such a monitoring system to be practical. Its cost should be limited, both by limiting changes to current hardware, and by minimizing the ongoing compliance costs to the Prover and enforcement costs to the Verifier.
The system should also not pose a high risk of leaking the Prover's proprietary information, including model weights, training data, or hyperparameters. 
Most importantly, the system must be robust to cheating attempts, even by highly-resourced adversaries such as government hacking groups, who may be willing to employ sophisticated hardware, software, and even supply-chain attacks.

\section{Solution Overview}\label{s.overview}

\begin{figure}
    \centering
    \includegraphics[width=\linewidth]{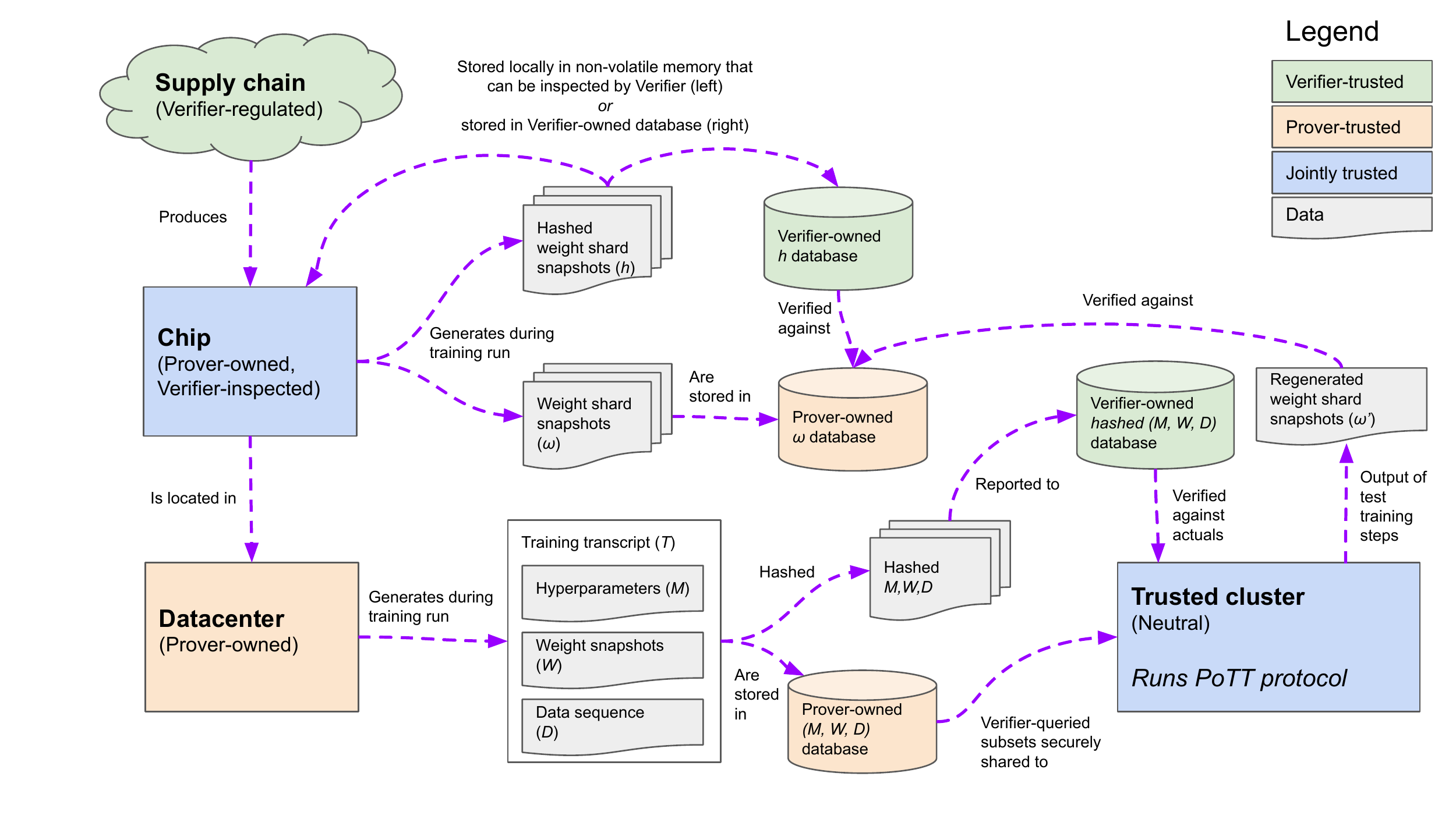}
    \caption{Overview of the proposed monitoring framework.}
    \label{fig:overview}
\end{figure}

In this section, we outline a high-level technical plan, illustrated in Figure \ref{fig:overview}, for Verifiers to monitor Provers' ML chips for evidence that a large rule-violating training occurred.
The framework revolves around chip inspections: the Verifier will inspect a sufficient random sample of the Prover's chips (Section \ref{s.sampling}), and confirm that none of these chips contributed to a rule-violating training run.
For the Verifier to ascertain compliance from simply inspecting a chip, we will need interventions at three stages: on the chip, at the Prover's data-center, and in the supply chain.
\begin{itemize}
    \item \emph{On the chip} (Section \ref{s.onchip}): When the Verifier gets access to a Prover's chip, they need to be able to confirm whether or not that chip was involved in a rule-violating training run.
    Given that rule violation depends only on the code that was run, our solution will necessitate that ML chips logging infrequent traces of their activity, with logging done via hardware-backed firmware.
    We suggest that ML chips' firmware occasionally log a copy of the current state of the chip's high-bandwidth memory to long-term storage, and in particular, that it logs the shard of the NN's weights stored in memory. 
    These \emph{weight-snapshots} can serve as a fingerprint of the NN training that took place on each chip.
    \item \emph{At the data-center} (Section \ref{s.datacenter}): The Verifier needs a way to interpret the chips' logs, and determine whether or not they are evidence for a rule-violating training run.
    To that end, the Prover, who is training the model, will be required to store a transcript of the training process --- including training data, hyperparameters, and intermediate weight checkpoints --- for each model they train.
    Using protocols similar to ``Proof-of-Learning'' \cite{jia2021proof}, these training transcripts may serve as provenance for the logged weight-snapshots, which are themselves the result of the same training process.
    In practice, for each (hash of a) weight-snapshot logged by a chip, the Prover provides the Verifier (the hashed version of) the matching training transcript.
    Then the Prover and Verifier jointly and securely verify that, with high probability, retraining using the training transcript would have indeed resulted in the logged weight-snapshot (and that no other valid training transcript could have resulted in that snapshot).
    Finally, now that the Verifier knows an approximate training transcript of the training run that had been executed on that chip at that time, they can examine properties of the training transcript to confirm that the Prover has complied with the agreed upon rules.
    \item \emph{At the supply chain} (Section \ref{s.supplychain}): The Verifier needs to know which ML chips the Prover owns, so that the Verifier can randomly inspect a representative sample of those chips, to confirm their ownership and that their logging mechanisms are reporting correctly.
    Without this chip-ownership verification step, a Prover might covertly acquire a large quantity of chips and use them for training without ever notifying the Verifier.
    Keeping track of chip-ownership is viable because the cutting-edge data-center chip supply chain is highly concentrated, meaning that chips originate from a few initial chokepoints and can be monitored thereafter.
\end{itemize}

These steps, put together, enable a chain of guarantees.
\begin{itemize}
\item
    When any organization wants to train a large rule-violating ML model, they must do so using chips that the Verifier is aware of.
\item
    These chips will occasionally log weight-snapshots.
    Each time a chip logs a weight-snapshot, the Prover must report the log to the Verifier, along with (hashes of) training transcripts that establish the provenance of that weight-snapshot.
\item
    If the Prover tries to hide the logs from every chip involved in the training run, the Verifier will eventually find out, because it will sample and physically inspect at least one of those chips with high probability.
\item
    Also, the hashed training transcripts that the Prover provides along with the logged weight-snapshot need to be authentic, reflecting the training run that was run on the chip.
    If they are not authentic, they will be caught by the Verifier's transcript verification procedure.
\item
    If the training transcript \emph{is} authentic, and the run violated the rules, then the Verifier can tell, and will catch the Prover.
\end{itemize}
    
Thus, so long as the Prover complies with the Verifier's steps, the Verifier will detect the Prover's rule-violation with high probability.
Just as in financial audits, a Prover's refusal to comply with the verification steps would itself represent an indication of guilt.

\subsection{Chip inspections}\label{s.inspection}
Ideally, chips could remotely report their logs, with on-chip firmware and remote attestation being sufficient to guarantee that those logs were truthfully reported (see Section \ref{s.onchip}).
Such guarantees may be sufficient to deter most Provers from misbehavior.
Unfortunately, these measures may not restrain well-resourced adversarial Provers:
it is generally understood within the hardware security world that, given unlimited physical access to a chip (which the Prover certainly would have), a chip's attestation and signed-firmware-enforcement mechanisms can be undermined \cite{skorobogatov2005semi}.
However, by physically inspecting chips after the fact, it may be possible to reliably detect such tampering.
This is because most hardware attacks require physically damaging the chip in ways that would be nearly-impossible to hide from a Verifier's subsequent inspection, especially if ML chips are supplemented with tamper-evident packaging measures \cite{waksman2010tamper}.
Luckily, rather than needing to inspect every chip, we only need to inspect enough to catch at least one chip from each large training run (Section \ref{s.sampling}).

A chip inspection can be done in one of two ways.
The Verifier can send an inspector to the facility in which a sampled chip is stored, which may be cost-effective if the facility contains multiple chips selected for sampling.
Alternatively, the ML chips can be \emph{transported a neutral inspection site}, as these chips are generally housed in easily-removable servers. (Sensitive components like storage can be removed prior to transport.)
The transporting option may be more attractive to security-conscious Provers who would prefer not to allow Verifier inspectors into their data-centers.

A chip inspection involves several steps.
First, the Verifier confirms the chip's serial number (often burned into chips and accessible both digitally and physically) matches the serial number they requested from the Prover.
This confirms that this is the correct randomly-sampled chip, and also that the Prover still owns the chip and has not secretly resold it to an untrusted party.
Second, the Verifier checks the on-chip logs, from which they extract the weight snapshot hashes which they will check against the Prover's reported training transcripts (Section \ref{s.datacenter}).
Finally, the Verifier checks that the chip's logging mechanisms, such as its secure firmware authentication mechanism (Section \ref{s.onchip}), have not been physically tampered with.

\subsection{How many ML chips does the Verifier need to inspect?}\label{s.sampling}
Sections \ref{s.onchip}, \ref{s.datacenter}, and \ref{s.supplychain} will together guarantee that if the Verifier observes even a single ML chip that logged a weight snapshot hash during a rules-violating training run, then the Verifier can use that weight snapshot to prove that the Prover violated the rules.
For the Verifier to make sure they'd observe at least one such snapshot from a large training run with high probability (or alternatively, catch a chip that'd been tampered with), they would need to inspect at least $s$ of the Prover's $\totalchips$ chips every $\monitoringperiod$ days.
We now seek to calculate $s$.

We define a large-scale training run as any training run using more than $\rulecomputeflops$ FLOPs.
We only seek to verify rules on training runs using $> \rulecomputeflops$ FLOPs, meaning that any weight-snapshots from the first $\rulecomputeflops$ FLOPs are not incriminating (as the Prover could have stopped the training run after that snapshot).
Catching a chip that happened to save a weight-snapshot at time $\rulecomputeflops + \epsilon$ is vanishingly unlikely for small $\epsilon$, so we instead provide guarantees on detecting the case where the Prover's training run is executed for $\geq 2 \rulecomputeflops$, substantially exceeding $\rulecomputeflops$.\footnote{We can always keep the detection threshold the same by cutting $\rulecomputeflops$ in half.
Also, the literature on neural scaling laws \cite{kaplan2020scaling, rosenfeld2021scaling, epoch2023scalinglawsliteraturereview} suggests that model loss decreases logarithmically at best given additional compute, so failing to catch a less-than-2x violation may have limited consequences.}
This means that the training run was executed for an additional second period of at least $\rulecomputeflops$ FLOPs, during which any weight snapshot would capture evidence of an ML model that had already been trained for at least $>\rulecomputeflops$ FLOPs.
From hereon, when we describe detecting a snapshot of a ``large-scale training run'', we refer to this second stage of the training run.
If a Verifier wanted to catch a Prover \emph{before} they had completed a $\rulecomputeflops$ rule-violating training run, they can simply shrink their target threshold to a fraction of $\rulecomputeflops$, and sample chips on a rolling basis.

The Verifier, whether due to specific suspicions or routine due diligence, wants to audit a particular Prover's $\totalchips$ chips.
Let $a$ be the FLOPs per day for those chips.
We will conservatively assume that NN training can be perfectly parallelized and utilize all the FLOPs on each chip.
Let $f$ be the expected number of weight-snapshots saved by the chip per day (Section \ref{s.onchip}) snapshots are Poisson distributed.
The Verifier wants to catch a training run within at most $\monitoringperiod$ days of its concluding, though the training run itself may have lasted for a longer period
\footnote{For simplicity, we assume $\trainingperiod$ is divisible by $\monitoringperiod$}
$\trainingperiod > \monitoringperiod$.
This implies that the Prover must have used a minimum
of $\frac{\rulecomputeflops}{a\trainingperiod}$ chips.
One might expect that by increasing $\trainingperiod$, the Prover could use fewer chips and therefore evade detection, but this turns out to not to meaningfully improve the Prover's evasion probability (a $<5\%$ increase) because each chip is sampled $\trainingperiod/\monitoringperiod$ times as often. (See Appendix \ref{a.provertricks} for details.)
Thus for simplicity, we assume the Verifier will choose $s$ as though $\trainingperiod$ is fixed.

To detect at least one chip with a snapshot from a rule-violating training run with probability $p$, the Prover needs to sample $s$ chips each monitoring period $\monitoringperiod$, where:
\begin{align}
\label{eq:samplingrate}
    s \geq \frac{\log{(1-p)}}{ \log{\left (1 - \frac{\rulecomputeflops}{a\trainingperiod \totalchips} \left ( 1 - e^{-f\monitoringperiod} \right ) \right ) }}
    \cdot \frac{\monitoringperiod}{\trainingperiod} 
\end{align}
We provide a derivation for Equation \ref{eq:samplingrate} in Appendix \ref{a.samplingderivation}.

In Table \ref{tab:samples}, we provide the total required samples-per-year $365 \cdot s/ \monitoringperiod$ to catch several past and projected training runs, for Provers who own different quantities of chips $\totalchips$.
\begin{table}[tbh]
    \begin{tabular}{
    l
    >{\raggedleft\arraybackslash}p{1.6cm}>{\raggedleft\arraybackslash}p{1.6cm}>{\raggedleft\arraybackslash}p{1.8cm}>{\raggedleft\arraybackslash}p{1.6cm}>{\raggedleft\arraybackslash}p{1.6cm}>{\raggedleft\arraybackslash} p{1.6cm}}
    \\
\textbf{Model}  & \textbf{Training FLOPs $\rulecomputeflops$}  & \textbf{H100-days $\rulecomputeflops/a$} & \textbf{H100s to train in 1 yr} & \textbf{Samples/yr if $C =10^3$} & \textbf{Samples/yr if $C =10^5$} & \textbf{Samples/yr if $C =10^7$} 
\\
\hline
    \\
GPT-3 \cite{BMR+20} & 3.14e+23 & 3.64e+3 & 10 & 243 & 2.43e+4 & 2.43e+6 \\
Chinchilla \cite{hoffman2022training} & 5.76e+23 & 6.67e+3 & 19 & 132 & 1.33e+4 & 1.33e+6 \\
PaLM \cite{chowdhery2022palm}& 2.56e+24 & 2.96e+4 & 82 & 29 & 2.98e+3 & 2.99e+5 \\
Chinchilla-280B $^\dag$  & 9.9e+24 & 1.15e+5 & 314 & 7 & 771 & 7.72e+4 \\
Chinchilla-1T $^\dag$ & 1.27e+26 & 1.47e+6 & 4.03e+3 & --- & 60 & 6.02e+3 \\
Chinchilla-10T $^\dag$ & 1.3e+28 & 1.5e+8 & 4.12e+5 & --- & --- & 58 \\
\\
    \end{tabular}
 \caption{
 \small{
 Example numbers of required total \emph{annual} samples $365 \cdot s / \monitoringperiod$ to catch a chip from every large-scale training run within $T=30$ days, 
 given $a = 10^{15} \cdot 24 \cdot 3600$ (the daily 16-bit Tensor Core FLOPs of an NVIDIA H100 SXM GPU \cite{nvidiah100}), 
 $f = 0.1$ weight snapshots per day (see Section \ref{s.onchip}), 
 and the Verifier's desired probability of catching a rule-violating training run $p=0.9$.
 Models marked with $\dag$ are projections for future training requirements \cite{hoffman2022training}.
 }
 }
\label{tab:samples}
\end{table}
As the table shows, the number of inspected chips remains approximately at a fixed fraction of the full stockpile, decreasing with training run size $\rulecomputeflops$.

An additional difficulty is the possibility that multiple Provers collude and each contribute a small fraction of their chips to a shared training run, such that if only one of the Provers is audited, the fraction of their $\totalchips$ chips they used for the training run is much smaller than $\frac{\rulecomputeflops}{a\monitoringperiod\totalchips}$.
This is particularly relevant in the International Oversight setting, where a government Prover might force the participation of multiple firms.
One solution to this is for an international Verifier to do background monitoring of \emph{all} Provers, in effect treating them all as one large Prover with access to the global chip stockpile.
Predictably, this would require a much greater number of samples, and is likely to only be feasible for regulating the largest-scale training runs (which can be caught with fewer samples).
Yet, for training runs at the scale of $10^{25}$ FLOPs or greater, monitoring could be done with a bureaucracy similar in size to the IAEA.\footnote{
We want to estimate the number of inspectors needed to catch a Chinchilla-280B-sized training run, with $10^{25}$ FLOPs, given several more years of hardware progress and global production.
Given $\totalchips = 10^7$ worldwide chips ($>5\times$ global stocks as of 2022), each of which can output $a=3\cdot 10^{15} \cdot 86400$ FLOPs per day ($3\times$ more FLOP/s than the NVIDIA H100), detecting a Chinchilla-280B-sized run within $T=30$ days of its completion anywhere on earth with 90\% probability would require roughly 232,000 worldwide chip samples per year.
A single inspector might be expected to verify at least 1000 chips a year, especially if those chips are brought to a central location (see Section \ref{s.inspection}).
This would require $\approx232$ inspectors, slightly smaller than the 280 active IAEA inspectors as of 2021. 
 \cite{dayinthelife}.}

As individual chips become more powerful ($a$ grows) and algorithmic efficiency increases ($\rulecomputeflops$ shrinks), the number of required samples would need to grow in order to verify rules on a fixed-capability training run.
Still, it may be that the training runs of greatest societal concern are those \emph{near the frontier} of scaling, and those are likely to remain detectable with few samples for the foreseeable future, even if Provers attempt to undermine them with physical tampering.



\section{On the chip}\label{s.onchip}
In an NN training cluster, a large number of ML devices (each containing an ML chip, and connected to a ``host'' CPU chip) collaborate to simultaneously compute gradient updates.
Large NNs have too many weights to fit in the high-bandwidth memory of a single ML device, so efficiently training a model requires splitting the model's weights
into \emph{weight shards}, and placing a shard on each chip \cite{smith2022computation}.
Each such shard is a vector of millions or billions of values.
Shards are updated after each gradient step so that new gradients can always be computed with respect to the latest weight values.
This means that at every timestep, the chip contains an up-to-date result from the training run, which we will show in Section \ref{s.datacenter} can serve as a type of checksum for the training run. \looseness=-1

In this section, we will describe how it may be possible to have ML chips log the values of these weight-shard vectors, for Verifiers to retroactively inspect.
We will assume the presence of certain hardware features\footnote{These include standard components of a hardware security module: firmware verification and attestation, firmware anti-rollback protection, and the ability to sign, authenticate, and encrypt messages.
It also includs the ability for firmware to raise interrupts, read device memory, and (to efficiently hash weight snapshots) execute code on the chip's processor.}
on the ML chip, though the details of the approach may need to be modified depending on the specifics of the device.
For example, if the ML chip does not have the hardware required for remote firmware attestation, the chip may need to be permanently mated to a host CPU that does have such hardware, with the CPU then orchestrating the logging operations.

A chip's weight-shards can be captured by \emph{snapshotting} the chip's high-bandwidth memory, which contains the weights.
These snapshots must be saved at random times, as otherwise the Prover can swap out the true weights for a fake set of weights right before the scheduled logging.
With some small probability at every clock cycle, the firmware present on each of the Prover's chips should raise an interrupt and briefly freeze the chip's operations.
The chip should then send a copy of the weight shard regions $\omega$ to its host CPU to save in Prover-controlled long-term storage.
The firmware also computes a hash $h = \hashfunction(\omega)$ of the relevant memory regions,
\footnote{Such hashing can be done cheaply so long as the firmware can leverage the ML chip's processing power.}
 using a publicly-agreed cryptographic hash function $\hashfunction$.
This hash $h$ is then saved for the Verifier to find it later, either by having the firmware store it on-chip\footnote{Keeping the hash in local NVRAM is useful if the ML chip is in a secure data-center and is restricted from communication with the outside world, such as via air-gapping.} in nonvolatile memory writable only by the firmware, or by having the chip sign the hash with its private key and then sending it to a Verifier-trusted central server (which must return an acknowledgement of receipt before the chip can resume operating).
Later, a Verifier can use $h$ to ask the Prover to provide them with the original weight shard $\omega$, as this is the only input to $\mathcal{H}$ whose hash matches $\hashfunction(\omega) = h$.

More frequent snapshots mean more performance overhead for the Prover, especially if all the other synchronized chips in the training run must hang until the single chip's snapshotting concludes. 
This freezing may become expensive if the rest of the training run must hang until it completes, especially if it's done by many chips.
\footnote{A possible alternative would be to orchestrate less-frequent snapshots via a Verifier-trusted central server, though this would require a solution to the problem of the Prover swapping out the chip's weights before propagating the Verifier's signal.}
The frequency of snapshots should be set as low as possible while still ensuring that used chips reliably contain a relevant weight snapshot if they were employed in a large-scale training run (see Eq. \eqref{eq:samplingrate}).

Though the Prover may claim to have used firmware that logged snapshots as described above, the Verifier may not trust the Prover to have indeed used the appropriate logging firmware.
For this reason, ML chips also need hardware mechanisms for enforcing that firmware be signed, and in turn, chipmakers will need to only sign firmware if it implements appropriate logging mechanisms.
\footnote{A similar firmware mechanism was used by NVIDIA to enforce cryptocurrency mining limits on its GPUs \cite{gartenberg_2021}.}
\footnote{Assuming that logging-free versions have been signed in the past, ML chips may need to include anti-rollback functionality \cite{krishnamurthy_odonoghue_bhatia}.
The Prover might also delay updating the chip's firmware until shortly before inspection.
To avoid this, ML chips may need to be made to remotely attest at an earlier time that their firmware has been updated.}
\footnote{Additional challenges arise if the Verifier suspects that the chipmaker (who authors the firmware) and Prover have colluded to create firmware with backdoors that disable logging.
Increasing Verifiers' confidence in the firmware may be an important consideration when verifying the operations of powerful nation-state Provers.}

An obstacle to logging all weight-shards stored in ML device memory is that different ML training code will store a model's weights in different regions of memory.
The chip's firmware must be able to determine which region of memory the weights are stored in.
It may be possible to determine the weight-regions retroactively, by logging the regions of memory claimed to correspond to the weights, along with a copy of the compiled on-device code, which can then be matched to Prover-provided source code and its memory allocation pattern analyzed.
\footnote{It may even be possible to modify standard libraries for generating chip-level ML training code (e.g., PyTorch-generated CUDA) to make their memory allocation processes more easily checkable by a subsequent Verifier.}
\footnote{Revealing the Prover's source code to the Verifier directly may be unacceptable, demanding a more complicated verification procedure like that described in Section \ref{s.realworld}.}
As a more invasive alternative, the Prover could proactively certify that its chip-level ML training code stores the weights in a specific memory region, by having its chip-code verified and then signed by a Verifier-trusted server before it can be executed by the firmware.
\footnote{The iOS App Store uses a similar method to ensure Apple devices only run signed programs \cite{codesigning}.}


A more challenging problem is that ``ML chips'' are general-purpose hardware accelerators, and thus are frequently used for other high-performance computing workloads like climate modeling.
There is no straightforward way to determine whether an ML chip is running a neural network training job (and therefore should log weight snapshots), or an unrelated type of job exempt from Verifier oversight.
\footnote{Potential avenues for addressing this may include be requiring non-ML-training code compilers to also sign their results, or improving methods for distinguishing between ML training code and other code.
If the types of code can be retroactively distinguished, then ML chips could all occasionally save memory/code snapshots, and then retroactively determine whether they belonged to a large-scale training run and thus deserve further scrutiny.
One particularly straightforward to address case is ML inference: the model's in-memory weights could be snapshotted and retroactively verified in a similar way to that described in Section \ref{s.datacenter}.
} 
Relatedly, it would also be desirable to be able to exempt a fraction of chips from monitoring, for use in education or small-scale R\&D where the overhead of compliance would be prohibitive.
\footnote{This might be addressable by having these ML chips' interconnect topology restricted to prevent their use in large-scale training.
However, methods for Verifiers to retroactively confirm the topology that ML chips were configured in at a data-center are beyond the scope of this work.}
Until we find methods for reliably distinguishing large-scale ML training, some fraction of ML chips must be exempted from using weight-logging firmware.
The larger system's guarantee would then be that \emph{for those of the Provers' chips that the Prover allows the Verifier to monitor}, no rule-violating training occurred.
The Verifier may then use their leverage to incentivize Provers into opting-in a larger and larger fraction of their chips over time.
\section{At the data-center}\label{s.datacenter}

From Sections \ref{s.sampling} and \ref{s.onchip}, we know how a Verifier can get access to at least one weight shard $\omega$ snapshotted during a large-scale training run.
In this section, we will show how the Verifier can use this weight shard to determine whether the associated training run followed the agreed-upon rules.

On its own, knowing $\omega$ is useless to the Verifier, since the Verifier does not know the values of the other weight-shards comprising the NN, or even how many other weight shards there are.
Yet $\omega$ holds the key to the training run's history. 
Any given weight-shard vector is incredibly unlikely to be yielded by the training process (there are $\gg 2^{10^6}$ possibilities even if we round to 1-bit precision).
But this particular $\omega$ is the nearly-deterministic result of the training run's hyperparameters $\mathbb{M}$ (including its random seed) and sequence of training data $\mathbb{D}$.
We refer to this information, along with information yielded throughout the training run $\mathbb{W}$ (such as intermediate weight snapshots), as the model's \emph{training transcript} $\mathbb{T} = \{ \mathbb{M}, \mathbb{D}, \mathbb{W}\}$.
It is plausible that it would be very difficult for the Prover to find a second ``legitimate'' training transcript $\mathbb{T}' = \{ \mathbb{M}', \mathbb{D}', \mathbb{W}' \} \neq \mathbb{T}$ that would exactly yield a full set of weights that included $\omega$ as a contiguous weight-shard.
It is also plausible that, if the Prover tried to lie and provide an $\mathbb{H}', \mathbb{D}'$, and $\mathbb{M}'$ that would \emph{not} result in $\omega$ but which they \emph{claimed} would result in $\omega$, then the Verifier might find a suite of tests that would detect such a faked transcript reliably.
If the Verifier could reliably detect such fakes, 
then by asking the Prover to provide a training transcript that would yield each logged snapshot $\omega$, the Verifier could force the Prover to reveal the true data $\mathbb{D}$, and hyperparameters $\mathbb{M}$ used in every large-scale training run (along with the full weights of the resulting model, not just $\omega$).
This would be sufficient information to check that the training run complied with any of the training rule types listed in Section \ref{s.rules}.

Unfortunately, having the Verifier simply re-run the full training process to verify a transcript's correctness is unworkable for several reasons.
First, the Prover would likely not be willing to reveal their training data, model weights, and hyperparameters, so the Verifier must do any verification without direct access to the data. (We address this in Section \ref{s.realworld}.)
Second, the compute cost to fully re-run the training transcript would be massive, as large as every original training run.
Third, the training run would likely not be perfectly reproducible: due to hardware-level noise, even two repetitions of the same sequence of training updates would gradually diverge.
Fourth, the Prover \emph{may} be able to construct a second ``spoof'' training transcript, that yields an exact match for $\omega$ but differs from the original training run that yielded $\omega$ in the first place.\footnote{As a trivial example, the Prover could claim that $\omega$ was simply the value of the random initialization, and no training had happened at the time of the snapshot.}

Thanfully, a close variant of this problem has already been studied in the literature, known as ``Proof of Learning'' \cite{jia2021proof}.
The goal of a Proof-of-Learning (PoL) schema is to establish proof of ownership over a model $W_t$ (e.g., to corroborate IP claims) by having the model-trainer save the training transcript $\mathbb{T}$ (including hyperparameters $\mathbb{M}$, data sequence $\mathbb{D}$, and a series of intermediate full-model weight checkpoints\footnote{We use ``weight checkpoints'' as shorthand, but if using an optimizer like Adam \cite{KingmaB14}, the optimizer state should also be included.} $\mathbb{W} = \{W_0, W_{k}, W_{2k} \dots \}$) which only the original model trainer would know.
Jia et al. \cite{jia2021proof} propose a verification procedure that makes it difficult for any third party to construct a spoofed transcript $\mathbb{T}'$, if they only have access to $W_t$ and the unordered dataset.

The solution of \cite{jia2021proof} is as follows: once a Prover reports a training transcript $\mathbb{T}$, the Verifier checks that the initialization appears random, and then chooses a number of pairs of adjacent weight snapshots that are $k$ gradient steps apart $(W_i, W_{i+k}), \dots, (W_j, W_{j+k})$.
Then, rather than re-running all of training, the Verifier only reruns the training of these specific segments, starting at $W_i$ and progressing through the appropriate data batches $D_i \dots D_{i+k}$ to yield a $W_{i+k}'$. 
The Verifier then confirms that the resulting checkpoint is approximately correct: $\| W_{i+k} - W_{i+k}'\|_2 < \epsilon$ for some small $\epsilon$ that depends on the reproduction noise and the number of steps $k$.
The original paper proposed a heuristic for choosing which segments to verify,
though subsequent papers have demonstrated spoofing attacks that evade the original verification approach \cite{zhang2022adversarial, fang2022fundamental}. 
Thus far, the literature on PoL has been heuristic-based, and discovering PoL protocols with higher degrees of reliability is the subject of ongoing work \cite{fang2022fundamental}.

\subsection{Proof-of-Training-Transcript}

Our requirement, which we will define as ``Proof-of-Training-Transcript'' (PoTT), differs from Proof-of-Learning because our Prover knows a true training transcript that leads to the final weight-snapshot (since they executed the original training run), and the Verifier wants to verify that the Prover has reported the \emph{original} transcript that yielded $\omega$ (or at least one very close to the original).
Further, unlike PoL, we allow that an adversarial Prover might be willing to spend \emph{more than the original training compute} $\rulecomputeflops$ in order to construct a spoof (e.g., an adversary may be willing to pay double the cost in order to execute a rule-violating training run without being detected).
Thus, we must hope to push the cost to construct a spoof to be as high as possible, so as to make evasion impractical.

\begin{definition}
 A PoTT protocol $(R, V)$ is defined by a Prover training and reporting protocol $R$ and a Verifier verification protocol $V$.
A target weight vector $W_t$ is produced as the true result of a training run which complies with a training and reporting protocol $R$, and which can be summarized via a training transcript $\mathbb{T} \in \mathcal{T}$, where $ \mathbb{T} = \{\mathbb{M}, \mathbb{D}, \mathbb{W}\}$, $\mathbb{M}$ is the metadata required to reproduce the training run (including hyperparameters, random seed, and loss function), $\mathbb{D}$ is an ordered sequence of sets of training data batches, and $\mathbb{W}$ is a collection of intermediate weight checkpoints resulting from the training procedure.
The verification protocol $V$ should accept any such true training transcript with high probability, $\Pr[V(\mathbb{T}, W_t) = \text{accept}] > 1 - \delta_1$ for some small $\delta_1$.

A ``spoofed'' training transcript $\mathbb{T}' = \{\mathbb{M}', \mathbb{D}', \mathbb{W}'\}$ is a transcript, which may not correspond to any valid training run, and which is substantially different from the original transcript $\mathbb{T}$ in its data or hyperparameters: $d_1(\mathbb{D}, \mathbb{D}') \geq \delta_3$ for some edit distance $d_1$ quantifying the number of data point insertions/deletions, and/or $d_2(\mathbb{M}, \mathbb{M}') \geq \delta_4$ for some hyperparameter distance $d_2$. 
A reporting/verification protocol pair $(R, V)$ is $J$-efficient and $F$-hard if $V$ runs in at most $J$ time, and there does not exist any spoof-generation algorithm $A \in \mathcal{A}: \mathcal{T} \rightarrow \mathcal{T}$ such that $\Pr[V(A(\mathbb{T}),W_t) = \text{accept}] > 1 - \delta_2$ where $A$ runs in less than $F$ time. \looseness=-1
\end{definition}

Colloquially, we want a Prover training and reporting protocol and Verifier verification protocol such that the Verifier only accepts \emph{original} training transcripts that would result in a final weight checkpoint which contains a shard matching our on-chip weight-shard snapshot $\omega$.
We leave the problem of finding provably secure, efficient methods for PoTT as an important avenue for future work, but we discuss a few promising directions below.

PoTT appears to be strictly harder than PoL, as it requires robustness to a better-resourced adversary has additional information (i.e., they know the true transcript $\mathbb{T}$) and has more compute-time to construct a spoof.
Given that existing PoL schemes are still heuristic-based and not yet provably secure, there may be a long way to go until we have PoTT methods that are both efficient and hard to spoof.
Still, one dynamic favoring the Verifier is that the Prover must \emph{commit} to a training transcript without knowing the Verifier's verification strategies.
Thus, Verifiers can amass secret collections of verification heuristics, much as the IAEA does not disclose all its methods for detecting nuclear activity.
Even if PoTTs are only ever heuristic-based, the presence of this dynamic may dissuade Provers from taking the risk of being detected by an unexpected test.

Defining conditions on the types of legitimate training runs is another useful source of leverage. 
For example, one Prover cheating strategy could be for the Prover to report one long training run as many shorter training runs, each claimed to be ``initialized'' where the previous training run ended.
A simple prevention would be for the training-and-reporting protocol $R$ to require the Prover to initialize every training run's weights via a known pseudorandom generator and a short seed.
This means that the randomness of the initial weights can later be confirmed by the Verifier.

Another promising strategy may be to require the Prover to \emph{pre-commit} to portions of its training transcript (e.g., the hyperparameters $\mathbb{M}$ and training batches $\mathbb{D}$) at the start of training.
This could be done by having the ML chip firmware log a hash of this precommitment, which would prove that the precommitment preceded the chip's snapshot $\omega$.
At the time of precommitment, the Prover does not know what trajectory the training run will follow or at what time it will be snapshotted, as the training has not yet been performed.
The Prover would be unable to construct a spoofed training transcript that would end at $\omega$ and precommit to it, because $\omega$ isn't known yet.
However, it is not obvious how to extend this approach to online learning settings like online RL, where the training data cannot be known ahead of time.

A final complication of our setting derives from the fact that the Verifier observes only a shard of the weights $\omega$, and not the full weight vector $W_t$.
It could be easier to construct a spoofed training transcript for some $\hat{W}$ which contains a shard matching $\omega$, but which differs from the true original weights $W_t \neq \hat{W}$ on the portion of the weight vector outside the shard.
We briefly describe an intuition for why this is likely to be as hard as general PoTT-spoofing.
Assuming $\omega$ must contain weights from more than a single linear layer, any Prover must at minimum construct a valid PoTT for this smaller NN represented by $\omega$, except without any original training transcript to start from (making it similarly hard to the original Proof of Learning problem).
Alternatively, if the Prover tries to reuse the original training transcript, it will require them to synthesize fake data and labels to exactly reproduce the true inputs/gradients, and these must correspond to a reasonable loss function that is actually decreasing over time.
If the Prover alternatively attempts to fake the surrounding layers' weights, then the weights in these surrounding layers must also have their own valid training trajectories as part of the training transcript \emph{and} must generate the appropriate inputs/gradients to yield $\omega$, simultaneously.

\subsection{Real-world implementation}\label{s.realworld}

As the literature uncovers new methods for verifying training transcripts, the Verifier can run these protocol to check that, for each of the logged weight snapshots on the Prover's chips that the Verifier inspects, the Prover has provided a legitimate corresponding training transcript.\footnote{
    Note that this requires the Prover to save and report training transcripts for all training runs corresponding to chips sampled by the Verifier, not just the largest-scale ones.
    This is because, without a matching training transcript, it's impossible for the Verifier to determine whether a given weight-shard was part of a large-scale training run or a small-scale one.
    Alternate methods for proving that a chip was only involved in a short training run are an important direction for future work.
    }
The Verifier can then examine the data, hyperparameters, and resulting models as reported by the trianing transcript, and thereby detect whether the Prover complied with the training rules.

A key design constraint is that the Prover does not disclose the raw transcripts (which likely contain confidential data that the Prover would prefer the Verifier did not see).
We propose a method for the Verifier to verify the Prover's previously-reported transcript's correctness without ever observing the raw data. 
First, the Prover and Verifier utilize a ``trusted cluster'': a set of chips that both parties trust, which is air-gapped to prevent the Prover's sensitive data from leaking, and which is equipped to execute small amounts of ML inference and training (so as to segment-wise approximately reproduce the Prover's original training process).
\footnote{Maintaining such compatible training clusters may prove quite challenging.
One desirable direction for future work is in verification methods by which the Verifier does not need to directly reexecute model training.
For example, it may be possible for the Verifier to interactively query the Prover for additional information on segments of the training run beyond what was included in the training transcript.
There may be queries that have valid answers \emph{only if} the original training transcript was authentic (e.g., a series of weight sub-checkpoints between two checkpoints, each with progressively lower loss), and the Prover could dynamically recompute answers to these queries using their own data-center.
While some properties of the verification would still need to be confirmed using a neutral cluster to maintain the confidentiality of the query-responses, such clusters may not need to be equipped for large-scale training, and thus be much easier to maintain.
}
Second, instead of reporting the full transcript, the Prover initially only provides the Verifier with hashes of the training transcript $\mathbb{T}$.
Then, at a later verification stage, the Prover supplies the true $\mathbb{M}, \mathbb{D}$, and $\mathbb{W}$ within the trusted cluster (which must match the reported hashes), executes the verification protocol, and reports whether both steps succeeded.
Finally, the Verifier can use this same trusted cluster to run programs checking that the Prover's training transcript complies with all agreed ML training rules (e.g., measuring the performance of the final model weights on agreed benchmarks).
We provide more details on a possible procedure in Appendix \ref{a.trustedcluster}. \looseness=-1

When evaluating which training transcript verification protocols to pursue, Verifiers should prioritize those verification strategies that get the most detection-probability, for the lowest cost.
Beyond the upfront costs of building trusted clusters or modifying chip hardware, the system has three ongoing operating costs: 
the efficiency loss from pausing to save weight checkpoints and the weight-shard snapshots (as described in Section \ref{s.onchip}), the storage costs for maintaining training transcripts (and in particular the weight-checkpoints, each of which may require terabytes) until the Verifier inspects them, and the compute costs to execute the training-transcript verification protocols on the trusted clusters.
These costs seem likely to scale linearly with the total compute used by the Prover, and will ultimately depend on the efficiency with which training transcripts can be verified.
Even though governments could in principle pressure Provers into paying the costs of compliance, a 1\% overhead for each dollar spent on training compute would be much easier for Provers to comply with than a 20\% overhead.
Indeed, for International Verifiers, the history of arms control suggests that maximally-stringent verification measures may have limited utility, as they may reduce the likelihood of compliance \cite{reddie2019governing}.
One important avenue for future work is finding cheaper, lower-fidelity alternatives to NN-retraining-based verification, which need only establish limited properties of the weight-shard's corresponding training run, and which could prompt more expensive verification methods if needed.
\section{At the supply chain}\label{s.supplychain}
We need supply-chain monitoring to accomplish two goals: to construct a ``chip directory'' of who owns each ML chip, for the purposes of sampling; and to ensure that each chip has the hardware features needed to provably log its training activity as in Section \ref{s.onchip}.
Unlike the chip and data-center interventions (Sections \ref{s.onchip} and \ref{s.datacenter}), monitoring the international ML chip supply chain cannot be done by a single Verifier.
Instead, an international consortium of governments may need to implement these interventions on behalf of other Verifiers (much as the IAEA runs inspections on behalf of member states).

\subsection{Creating a chip-owner directory}\label{s.chipowner}
For a Verifier to be confident that a Prover is reporting the activity of all the Prover's ML chips, they need to know both which ML chips the Prover owns, and that there are no secret stockpiles of chips beyond the Verifier's knowledge.
Such ownership monitoring would represent a natural extension of existing supply chain management practices, such as those used to enforce U.S. export controls on ML chips.
It may be relatively straightforward to reliably determine the total number of cutting-edge ML chips produced worldwide, by monitoring the production lines at high-end chip fabrication facilities.
The modern high-end chip fabrication supply chain is extremely concentrated, and as of 2023 there are fewer than two dozen facilities worldwide capable of producing chips at a node size of 14nm or lower \cite{semiconductorwiki}, the size used for efficient ML training chips.
As \cite{baker2023nuclear} shows, the high-end chip production process may be monitorable using a similar approach to the oversight of nuclear fuel production (e.g., continuous video monitoring of key machines).

As long as each country's new fab can be detected by other countries (e.g., by monitoring the supply chain of lithography equipment), an international monitoring consortium can require the implementation of verification measures at each fab, to provide assurances for all Verifiers.
After processing, each wafer produced at a fab is then sent onward for dicing and packaging.
Since the facilities required for postprocessing wafers are less concentrated, it is important for the wafers (and later the dies) to be securely and verifiably transported at each step.
If these chip precursors ever go missing, responsibility for the violation would lie with the most recent holder.
This chain of custody continues until the chip reaches its final owner, at which point the chip's unique ID is associated with that owner in a \emph{chip owner directory} trusted by all potential Verifiers and Provers.
This ownership directory must thereafter be kept up-to-date, e.g., when chips are resold or damaged.\footnote{In the rare scenario where a large number of chips owned by the same Prover are lost or destroyed beyond recognition, the Verifier or international consortium can launch an investigation to determine whether the Prover is lying to evade oversight.}
The continued accuracy of this registry can be validated as part of the same random sampling procedure discussed in Section \ref{s.inspection}.
As a second layer of assurance, chips could also be discovered by inspecting datacenters, if those datacenters are detectable via other signals \cite{baker2023nuclear}.

Given the high prices and large power and cooling requirements of these ML chips, they are largely purchased by data-center operators.
These organizations are well-suited to tracking and reporting transfers of their ML chips, and complying with occasional inspections. 
Though a small fraction of data-center ML chip purchases are made by individuals, so long as these are a small fraction of chips they may be exempted from the overall monitoring framework.

\subsection{Trusting secure hardware}
We require in Section \ref{s.onchip} that each ML chip produced by the semiconductor supply chain is able to provably log traces of its usage.
The second goal of supply-chain monitoring is to provide Verifiers with high confidence in the reliability of these on-chip activity-logging mechanisms.
This requires ML chip designers to integrate security features into their hardware and firmware designs, especially in ways that make them externally-legible to Verifiers that may not trust the chip-designer.
Key priorities include the immutability of the chip's burned-in ID, the integrity of the hardware-backed mechanism for only booting signed firmware, and the resilience of the on-chip hardware-roots-of-trust to side-channel attacks that could steal the chip's encryption keys \cite{crypto-1996-1469, 6271612} and thus fake its logs.

A concern for Verifiers checking the conduct of powerful Provers (e.g., states verifying each others' ML training runs) is the possibility of supply-chain attacks \cite{robertson_riley_2021}, which could enable a Prover to undetectably disable/spoof the ML chips' logging functionality.
Fully mitigating the threat of supply-chain attacks is a major global issue and beyond the scope of this paper.
However, one particularly useful step for building trust in ML chip mechanisms' integrity would be for ML chip designers to use open-source Hardware-Roots-of-Trust.
This transparency means that chips' designs can be validated by untrusting Verifiers to confirm there are no backdoors.
For example, Google's Project OpenTitan has produced such an HRoT \cite{opentitan}, and many major ML chip designers (Google, Microsoft, NVIDIA, and AMD) have agreed to integrate the Open Compute Project's ``Caliptra'' Root of Trust. \cite{caliptra}

\section{Discussion}\label{s.discussion}
The described additions to the production and operation of ML training chips, if successfully implemented, would enable untrusting parties (like a government and its domestic companies, or the US and Chinese governments) to verify rules and commitments on advanced ML development using these chips.
There are many useful measures that governments and companies could begin taking today to enable future implementation of such a framework if it proved necessary, and that would simultaneously further businesses' and regulators' other objectives.
\begin{itemize}
    \item Chipmakers can include improved hardware security features in their data-center ML chips, as many of these are already hardware security best practices (and may already be present in some ML chips \cite{nvidiah100}).
    These features are likely to be independently in-demand as the costs of model training increase, and the risk of model theft becomes a major consideration for companies or governments debating whether to train an expensive model that might simply be stolen.
    \item Similarly, many of the security measures required for this system (firmware and code attestation, encryption/decryption modules, verification of produced models without disclosing training code) would also be useful for ``cloud ML training providers'', who wish to prove to security-conscious clients that the clients' data did not leave the chips, and that the clients' models did not have backdoors inserted by a third party \cite{liu2017trojaning}.
    Procurement programs like the US's FedRAMP could encourage such standards for government contracts, and thereby incentivize cloud providers and chipmakers to build out technical infrastructure that could later be repurposed for oversight.
    \item Individual companies and governments can publicly commit to rules on ML development that they would like to abide by, if only they could have confidence that their competitors would follow suit.
    \item Responsible companies can log and publicly disclose (hashed) training transcripts for their large training runs, and assist other companies in verifying these transcripts using simple heuristic.
    This would not prove the companies \emph{hadn't also} trained undisclosed models, but the process would prove technical feasibility and create momentum around an industry standard for (secure) training run disclosure.
    \item Companies and governments can build trusted neutral clusters of the sort described in Section \ref{s.realworld}.
    These would be useful for many other regulatory priorities, such as enabling third-party auditors to analyze companies' models without leaking the model weights.
    \footnote{For similar reasons, the US Census Bureau operates secured Federal Statistical Research Data Centers to securely provide researchers access to sensitive data \cite{uscensusbureau_2022}.}
    \item Governments can improve tracking of ML chip flows via supply-chain monitoring, to identify end-users who own significant quantities of ML chips.
    In the West, such supply-chain oversight is already likely to be a necessary measure for enforcing US-allied export controls.
    \item Responsible companies can work with nonprofits and government bodies to practice the physical inspection of ML chips in datacenters.
    This could help stakeholders create best practices for inspections and gain experience implementing them, while improving estimates of implementation costs.
    \item Researchers can investigate more efficient and robust methods for detecting spoofed training transcripts, which may be useful for in proving that no backdoors were inserted into ML models.
\end{itemize}
For the hardware interventions, the sooner such measures are put in place, the more ML chips they can apply to, and the more useful any verification framework will be.
Starting on these measures early will also allow more cycles to catch any security vulnerabilities in the software and hardware, which often require multiple iterations to get right.

\subsection{Politics of Implementation}
Given the substantial complexity and cost of a monitoring and verification regime for large-scale ML training runs, it will only become a reality if it benefits the key stakeholders required to implement it.
In this last section, we discuss the benefits of this proposal among each of the required stakeholders.
\begin{itemize}
    \item \emph{The global public}: Ordinary citizens should worry about the concentration of power associated with private companies possessing large quantities of ML chips, without any meaningful oversight by the public.
    Training run monitoring is a way to make powerful companies' advanced ML development accountable to the public, and not just the free market.
    Most importantly, ordinary people benefit from the security and stability enabled by laws and agreements that limit the most harmful applications of large-scale ML systems.
    \item \emph{Chipmakers and cloud providers}: Absent mechanisms for verifying whether ML chips are used for rule-violating training runs, governments may increasingly resort to banning the sale of chips (or even cloud-computing access to those chips) to untrusted actors \cite{bis2022controls}.
    By enabling provable monitoring of large-scale ML training runs, chipmakers may reverse this trend and may even be able to resume sales to affected markets.
    \item \emph{AI companies}: Responsible AI companies may themselves prefer not to develop a particular capability into their products, but may feel they have no choice due to competitive pressure exerted by less-scrupulous rivals.
    Verifying training runs would allow responsible AI companies to be recognized for the limits they impose on themselves, and would facilitate industry-wide enforcement of best practices on responsible ML development.
    \item \emph{Governments and militaries}: Governments' and militaries' overarching objective is to ensure the security and prosperity of their country.
    The inability to coordinate with rivals on limits to the development of highly-capable ML systems is a threat to their own national security.
    There would be massive benefit to a system that enabled (even a subset of) countries to verify each others' adherence with ML training agreements, and thus to maintain an equilibrium of responsible ML development. 
\end{itemize}
Even if only a subset of responsible companies and governments comply with the framework, they still benefit from verifiably demonstrating their compliance with self-imposed rules by increasing their rivals' and allies' confidence in their behavior \cite{horowitz2021ai} (and thus reducing their rival's uncertainty and incentive towards recklessness).

Finally, we highlight that the discussed verification framework requires continuous participation and consent by the Prover.
This makes the framework fundamentally non-coercive, and respects national sovereignty much as nuclear nonproliferation and arms control agreements respect national sovereignty.
Indeed, the ongoing success of such a system relies on all parties' self-interest in continuing to live in a world where no one -- neither they, nor their rivals -- violates agreed guardrails on advanced ML development.

\section*{Acknowledgements}
The author would like to thank 
Tim Fist,
Miles Brundage,
William Moses,
Gabriel Kaptchuk,
Cynthia Dwork,
Lennart Heim,
Shahar Avin,
Mauricio Baker,
Jacob Austin,
Lucy Lim,
Andy Jones,
Cullen O'Keefe,
Helen Toner,
Julian Hazell,
Richard Ngo,
Jade Leung,
Jess Whittlestone,
Ariel Procaccia,
Jordan Schneider,
and Rachel Cummings Shavit
for their helpful feedback and advice in the writing of this work.

\printbibliography

\appendix

\newpage

\section{Discussion on future training requirements}\label{app.howmuch}

\subsection{Will the most capable ML models require large-scale training?}
This paper's proposed framework is premised on the assumption that large-scale training is and continues to be a necessary requirement for the most advanced (and thus most dangerous) ML models.
There is intense disagreement within the field about how important large-scale training is, and how long that will remain the case.

Many of the recent breakthroughs in machine learning model capabilities, across every domain, have come from increasing the model size or quantity of training data, each of which corresponds to a greater usage of compute \cite{kaplan2020scaling, hoffman2022training, zhai2021scaling}.
Indeed, some capabilities, such as chain-of-thought reasoning, appear to only emerge at the largest training scales \cite{wei2022chain}.
At the same time, any one narrow capability can often be achieved with a much smaller compute budget \cite{magister2022teaching, madani2023large}.
Nonetheless, Sutton's ``Bitter Lesson'' \cite{sutton2019bitter} that ``general methods that leverage computation are ultimately the most effective'' is a frequent diagnosis of the likely future of deep learning.
Though algorithmic progress \cite{erdil2022algorithmic} and the continued progress of Moore's Law will continue to reduce the number of chips required for any specific capability, we may compensate by gradually increasing enforcement parameters to work for smaller quantities of specialized compute.
At the same time, the increasing investment in compute by frontier AI firms \cite{lardinois_2022, wiggers_2022} suggests that industry insiders continue to believe that the most capable frontier models --- likeliest to yield new capabilities and surface new risks to public safety --- are expected to require ever more compute.

\subsection{Will large-scale training continue to require specialized datacenter chips?}

Nearly all large-scale training runs are executed on high-end datacenter accelerators \cite{chowdhery2022palm, kaplan2020scaling, zeng2022glm}.
The main difference between these chips and their consumer-oriented counterparts is their much higher inter-chip communication bandwidth (e.g., 900GB/s for the NVIDIA H100 SXM vs. 64GB/s for the NVIDIA GeForce RTX 4090 \cite{nvidiah100, nvidia4090}).
This extra bandwidth is today crucial for parallelizing NN training, especially tensor parallelism and data parallelism, which require frequent transfers of large matrices between many chips \cite{smith2022computation}.
Organizations doing large-scale training also favor these datacenter chips for other reasons: they are generally more energy efficient, and license requirements often prevent organizations from placing consumer-oriented chips in datacenters\cite{moss_2023}.

Still, recent work has suggested it may be \emph{possible} to do large-scale training on consumer chips with low interconnect, though with substantial cost and speed penalties\cite{binhang2022distributed, ryabinin2023distributed}.
If such methods become feasible for bad actors, then we may need to adjust to a different regulatory model for detecting training activity.
Possibilities include focusing on spotting and monitoring datacenters (similar to the IAEA's work to detect undeclared nuclear facilities \cite{harry1996iaea}), or regulating the high-capacity switches that could be necessary to enable fast networking between low-interconnect chips.
So long as they can be detected, it may be possible to retrofit consumer chips (e.g. with a permanently-mated host CPU, see Section \ref{s.onchip}) to enable similar monitoring capabilities.

It is important to note that the current framework \emph{does} apply in the setting where clusters of chips are split across several datacenters (e.g. multiple cloud providers), so long as these high-end chips are used at each datacenter.

\section{Derivation of Sampling Rate}
\label{a.samplingderivation}
We provide a derivation of Equation \ref{eq:samplingrate}, the number of samples required for a Verifier to catch a weight snapshot from a rule-violating training run with a given probability $p$.
Let $\rulecomputeflops$ be the size of an ML training run that the Verifier is hoping to catch.
Let $\totalchips$ be the total number of chips the Prover possesses, and $a$ be the FLOPs/day for those chips. 
Let $f$ be the expected number of weight-snapshots saved by the chip per day; snapshots are Poisson distributed.
The Verifier wants to detect a rule-violating training run of length $\trainingperiod$ that was completed in the last $\monitoringperiod$ days, and will sample $s$ chips every $\monitoringperiod$.
We will assume that the Prover executes the training run over the course of $\trainingperiod$ days. 
(We show in Appendix \ref{a.provertricks} that changing $\trainingperiod$ only marginally affects our analysis.)

It is not enough that a chip involved in the training run be sampled by the Verifier; the chip needs to have also logged a weight-snapshot from this particular training run, in order for there to be something for the Verifier to discover.
The probability of a chip logging a weight-snapshot is uniform over time (Section \ref{s.onchip}). 
Assuming the training run began at the beginning of the first monitoring period of length $\monitoringperiod$, the probability that a snapshot was sampled is $1 - PoissonCDF_{T;f}(0) = 1 - e^{-fT}$, where $PoissonCDF_{T;f}$ is the CDF of a Poisson random variable with shape parameter $f$ and interval-length $T$.
If the training run lasted multiple monitoring periods, then each later period, each chip is strictly more likely to contain a snapshot than this first period.
\footnote{There are two edge cases.
First, the Prover could choose to use extra chips and thus shrink $\trainingperiod < \monitoringperiod$.
However, in Appendix \ref{a.trainingshorter}, we show this would not improve the likelihood of avoiding detection due to a snapshot not being included.
The other edge case is when $\trainingperiod$ is not perfectly divisible by $\monitoringperiod$, leading to the first round of samples occurring when each chip has participated in the training run for less than $\monitoringperiod$ time.
This means that the likelihood of each sample in that round containing a snapshot is slightly reduced, thus reducing the likelihood of detection in that round, and therefore of detection at any of the $\trainingperiod/\monitoringperiod$ periods of the training run.
At worst, this creates a delay of one extra monitoring period $\monitoringperiod$ after the training run ended, since each sample from that $\lceil \trainingperiod / \monitoringperiod \rceil$'th period is just as likely to contain a snapshot as the other periods.
That's because the weight snapshots remain on the chip even after the training run has ended.
We exclude this one-period fudge factor from our notation for brevity.}

Thus, we let the probability that a chip contains a snapshot be:

\[
p_w \geq 1 - e^{-fT}
\]

We assume that the Prover used the minimum number of chips possible to complete the training run within the time period, $\frac{\rulecomputeflops}{a\trainingperiod}$ chips.
Each time a Verifier samples a chip, the probability that it is one of the chips used in the training run is $\frac{\rulecomputeflops}{a \trainingperiod\totalchips}$.
The probability that the Verifier samples a chip, and that it is both used in the training run and contains a logged weight snapshot is $p_{\text{1-sample}} = \frac{\rulecomputeflops}{a\trainingperiod\totalchips} p_w $.

The probability that the Verifier \emph{fails} to sample a single involved chip with a logged snapshot during a single monitoring period that includes $s$ samples is equivalent to a weighted coin with $p_{\text{1-sample}}$ probability of heads coming up tails $s$ times:
\[
p_n = (1 - p_{\text{1-sample}})^s = \left( 1 - \frac{\rulecomputeflops}{aT\totalchips}\left(1 - e^{-fT} \right)\right)^s
\]

There are $\lceil\trainingperiod/\monitoringperiod\rceil$ rounds of sampling for each training run, including the last period (which may not include a full training run, but each chip will still store the relevant weight snapshot hashes).

This means the probability of successfully detecting a snapshot from a large-scale training run within $\monitoringperiod$ days of the training run's completion is $p = (1 - p_n)^{\lceil \trainingperiod/\monitoringperiod \rceil}$.

Rearranging terms, we get:
\begin{align*}
p &= 1 -\left( 1 - \frac{\rulecomputeflops}{a\trainingperiod\totalchips} p_w \right)^{\lceil \trainingperiod / \monitoringperiod \rceil s}
\geq 
1 -\left( 1 - \frac{\rulecomputeflops}{a\trainingperiod\totalchips}\left(1 - e^{-f\monitoringperiod} \right)\right)^ {\lceil \trainingperiod / \monitoringperiod \rceil s} \\
\log(1 - p) &\leq s 
\left \lceil \frac{\trainingperiod}{\monitoringperiod} \right \rceil 
\log\left( 1 - \frac{\rulecomputeflops}{a\trainingperiod\totalchips}\left(1 - e^{-f\monitoringperiod} \right)\right) \\
s &\leq \frac{ \log (1-p)}{\log\left( 1 - \frac{\rulecomputeflops}{a\trainingperiod\totalchips}\left(1 - e^{-f\monitoringperiod} \right)\right)}
\frac{1}{\left \lceil \frac{\trainingperiod}{\monitoringperiod} \right \rceil}
\end{align*}
To catch a rule-violating training run snapshot with probability $p$, a Verifier needs to sample no more than $s$ chips in each monitoring period $\monitoringperiod$.
\section{Analysis of prover evasion strategies}
\label{a.provertricks}
\subsection{Do the Prover's odds of evasion benefit from training for longer, across fewer chips?}
\label{a.traininglonger}
One concern is that a Prover might manage to evade detection of a large-scale training run by using a smaller number of chips over a long period of time. 
A hypothesis for why this might be beneficial is that using a smaller number of chips makes the likelihood of any one of them being sampled extremely small.
However, if the Prover uses $\frac{c}{k}$ chips, this increases the training run's length $\times k$, meaning that the number of rounds of sampling similarly increases $\times k$.
We explore which of these effects dominates.

This can be expressed in the following formula for $p_d(k)$, the probability of detection as a function of the multiple increase of the training length. (For simplicity, we assume that the snapshotting frequency $f$ is high enough that every sampled chip contains a snapshot.)
\[
p_d(k) = 1 - \left ( 1 - \frac{\rulecomputeflops}{\totalchips a \trainingperiod k}\right )^{ks}
\]
where $a$ is the chips' FLOPs per day, $\trainingperiod$ is the original training period, $\totalchips$ is the total size of the Prover's chip stockpile being sampled from, and $\rulecomputeflops$ is the quantity of compute required for a large-scale training run.

We assume that $s$ is set by the Verifier in order to achieve a particular detection probability $p$ assuming the original (un-extended) training period $T$, via the Equation \ref{eq:samplingrate}, reproduced here:
\[
s(p) = \frac{\log (1-p)}{\log \left (1 - \frac{\rulecomputeflops}{\totalchips a \trainingperiod}\right )}
\]

\begin{figure}
    \centering
    \includegraphics[width=0.5\linewidth]{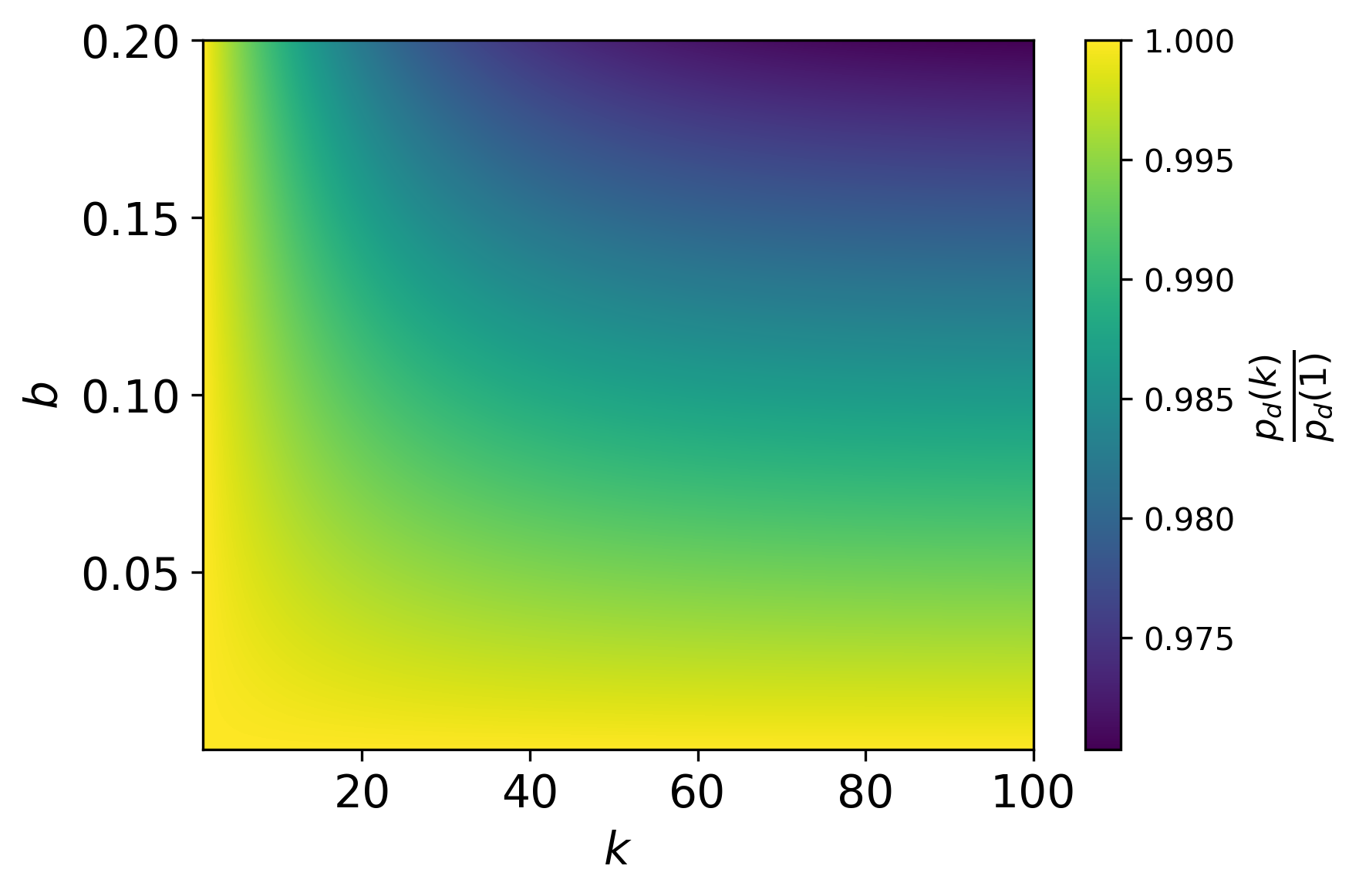}
    \caption{Ratio of $p_d(k)$ to original $p = p_d(1)$, setting $p=0.9$ and varying $k$ from $1$ to $100$ and $b = \rulecomputeflops/(\totalchips a \trainingperiod)$ from $0$ to $0.2$ (where the latter end corresponds to the Prover using 20\% of all chips, making such a large-scale training run easily catchable with even a dozen samples). Across all values, the Prover's reduction in probability of detection is less than 5\%, even when training for up to $100 \times$ longer.
    }
    \label{fig:longertimes}
\end{figure}

In Figure \ref{fig:longertimes}, we estimate how $p_d(k)$, the probability of detection given a $\times k$ longer run, changes with $k$ and other hyperparameters. For simplicity, we define $b = \rulecomputeflops/(\totalchips a T)$, where we know that $1>b>0$.

Somewhat surprisingly, $p_d(k)$ changes very little as $k$ increases, and this remains true for almost all values of $b$.
This means that, for nearly all training run sizes (affecting $b$) and desired Verifier confidence levels $p$, the Prover cannot meaningfully decrease their probability of having their rule-violating training run detected by the Verifier by simply using fewer chips for longer.

\subsection{Do the Prover's odds of evasion benefit from splitting training across more chips?}
\label{a.trainingshorter}

A Prover may also exploit each chip's low snapshotting frequency to train across more chips than necessary, with the knowledge that many of the used chips will not contain a single snapshot.
Thus, the Verifier might sample more chips that were involved in a rule-violating training run, but never detect their involvement.

However, since the total amount of compute-time stays constant, the total number of snapshots stays constant, and using more chips only \emph{increases} the probability of detection.
We formalize this intuition with the following proof.

\begin{theorem}
For a given firmware snapshotting frequency $f$ and maximum training time $\trainingperiod$, a Prover executing a rule-violating training run cannot reduce the probability of one of the training run's snapshots being detected (that is, of having at least one of the Verifier's sampled chips include a snapshot from the rule-violating training run) by splitting the training run across a greater number of chips, each processing for less time.
\end{theorem}

\begin{proof}

We prove this in the case where training is done for at most one monitoring period $\trainingperiod = \monitoringperiod$.
The extension to multiple monitoring periods proceeds automatically.

Let the total compute required for a rule-violating training run be $\rulecomputeflops$ FLOPs, and the total number of chips available to the Prover is $\totalchips$.
Given that each chip can process $a$ FLOPs per day, to complete the training run in $t$ time, we need to use $c(t) = \rulecomputeflops/(at)$ chips.

For a chip used in the training run for $t$ time, the probability $p_{ns}$ that no snapshot was saved is the CDF of a Poisson random variable with rate parameter $f$:
\[
p_{ns}(t) = e^{-ft}
\]
Let the Verifier's sampling rate be a total of $s$ chips, which we assume occurred after the end of the training run (to avoid complexity due to chip samples midway through the training run having a lower probability of having logged a snapshot than later samples in the same training run).

If $c(t)$ chips are used, each for $t$ time, then the overall probability of detection $p_d(t)$ is
\[
p_d(t) = 1 - \left (1 - \frac{c(t)}{\totalchips} \left (1 - p_{ns}(t) \right ) \right )^s = 1 - \left (1 - \frac{\rulecomputeflops}{a\totalchips t} \left (1 - e^{-ft} \right ) \right )^s
\]
We want to prove that if the training run uses more chips than necessary $c(t) > c(T)$, each for less time $t < T$, then detection is always more likely $p_d(t) \overset{?}{>} p_d(T)$.

Let $a = \frac{T}{t} \geq 1$, and let $b = e^{-fT}$. We know $1>b>0$ because $f$ and $T$ are positive.

\begin{lemma}
If $g \geq 1$ and $1>b>0$, then $f(a) = a \left (1-b^{\frac{1}{g}} \right ) - (1-b) \geq 0$.
\end{lemma}
\begin{proof}
\begin{align*}
f(0) &= (1-b) - (1-b) = 0 \\
\frac{df(g)}{dg} &= 1 - b^{\frac{1}{g}} + \frac{1}{g}b^{\frac{1}{g}} \\
&= 1 - \underbrace{\left (1 - \frac{1}{g} \right )}_{\leq 1}\underbrace{b^{\frac{1}{g}}}_{\leq 1} \\
&\geq 0
\end{align*}
Given that $f(g=0) \geq 0$ and its derivative is continuous and nonnegative thereafter, we know that $f(g) \geq 0$ for all $g \geq 1, 1>b>0$.
\end{proof}
Plugging in our earlier settings for $a$ and $b$, we have:
\begin{align*}
   g\left (1 - b^{\frac{1}{g}} \right ) - (1-b) &\geq 0 \\
   g\left (1 - b^{\frac{1}{g}} \right ) &\geq 1-b \\
   \frac{1}{t}\left(1 - e^{-ft}\right) &\geq \frac{1}{T}\left(1 - e^{-fT}\right ) \\
   \frac{\rulecomputeflops}{a \totalchips t}\left (1 - e^{-ft}\right ) &\geq
   \frac{\rulecomputeflops}{a \totalchips T}\left (1 - e^{-fT} \right ) \\
   1 - \left ( 1 - \frac{\rulecomputeflops}{a \totalchips t}\left (1 - e^{-ft} \right ) \right )^s &\geq 
   1 - \left ( 1 - \frac{\rulecomputeflops}{a \totalchips T}\left (1 - e^{-fT} \right ) \right )^s\\
   p_d(t) &\geq p_d(T)
\end{align*}
\end{proof}

Note that this proof applies whether the extra chips are being used simultaneously in parallel, or split across time (e.g. one group of chips for the first half of the training run, and a second group of chips for the second half).

While this result might seem to contradict the result from Subsection \ref{a.traininglonger}, the two findings are in fact consistent. The key assumption is that the snapshotting frequency $f$ is high enough for the chips to contain at least one logged snapshot per period $\monitoringperiod$ with high probability.
Then Subsection \ref{a.trainingshorter} tells us that using fewer chips reduces the Prover's probability of detection, and Subsection \ref{a.traininglonger} tells us that this decrease in detection is small, and likely not worth it to the Prover given the resulting delay in the training run's completion.
\section{Logistics for Confidentially Verifying Proofs-of-Training-Transcript}
\label{a.trustedcluster}

In this section, we consider the logistical problem of how to verify, for all the Verifier's observed weight-snapshots from sampled chips, that the Prover has provided authentic training transcript that match each weight-shard. We assume the PoTT verification protocol looks similar to the  Proof-of-Learning scheme of \cite{jia2021proof}.
Our solution is specifically designed to avoid the Prover ever having to directly reveal the hyperparameters $\mathbb{M}$, training data $\mathbb{D}$, and model weights $\mathbb{W}$ to the Verifier, as in many cases these data are either private or proprietary.

First, when executing each training run, the Prover follows any required\footnote{Of course, these restrictions must be retroactively verifiable using the training transcript, as otherwise the Prover might simply not comply.}
restrictions on their training (e.g., starting from a subset of valid initializations, or precommitting to the model weights).
Throughout training, the Prover saves the hyperparameters (e.g., training code), the sequence of training data batches, and periodic snapshots of the model weights.
They also make sure to track any randomness used, and generally make sure their training process is repeatable (up to low-level noise) assuming access to similar but non-identical hardware configurations.
This itself is a nontrivial technical challenge \cite{liu2021reproducibility}, especially in co-training settings like GANs or RL on learned reward models, and would benefit from further investments in replication tools for common ML frameworks.

Periodically, the Prover reports their chips' on-chip logged weight-shard snapshot hashes, along with hashed versions of the Proofs-of-Training-Transcript matching each of these logged hashes (including hashes of $\mathbb{M}$, hashes of each model snapshot $\mathbb{W}$, and sequences of hashes of training data points $\mathbb{D}$).
The Prover may need to further disclose basic information like the distance between weight-snapshot pairs, which may be needed for the Verifier to determine which regions of the transcript to verify.\footnote{This info could be proven to the Verifier securely and privately, for example by using standard ZK-SNARK proof tools to confirm that two given hashes correspond to vectors that have a given $L_2$ distance between them.}

As in Section \ref{s.sampling}, the Verifier chooses a subset of these chips whose logs will be confirmed via physical inspection.
The Verifier inspects the training transcripts and associated metadata, and chooses the subsets of the training transcripts to verify.
They also select any other components of the training transcript (e.g., a random sample of the training data) that they will need in order to determine that that the training run has complied with the ML training rules.
At this point, the Verifier has a set of hashes of the inputs to its verification protocol (including model weight snapshots and training data points), but does not have access to any of this information directly.

The Verifier needs to have faith in the results of the computation, but without seeing the inputs themselves.
In principle, this might be addressed by cryptographic techniques like zero-knowledge proofs, but in practice these techniques cannot efficiently execute computationally intensive programs, like long sequences of gradient updates on billion-parameter models.
Instead, the Prover and Verifier can agree on a neutral jointly-trusted cluster, which they will use to execute the verification protocol.
The Verifier needs to be able to trust in the integrity of the cluster's computation.
The Prover, conversely, needs to trust that the cluster will not reveal its private data, by verifying that it cannot communicate with the outside world and has no persistent storage.
Such neutral clusters could be maintained by a trusted third party, or when no such trusted party exists, could be jointly overseen by both the Prover and Verifier. 
It may be useful to create one trusted cluster for each type of ML training chip, in order to mitigate potential hardware compatibility issues.

At agreed-upon intervals, the Verifier supplies the hashes of the inputs (including hyperparameters $\mathbb{M}$) to the neutral cluster, and the Prover supplies the inputs themselves.
The cluster hashes the Prover's inputs and confirms they match the hashes.
Then the cluster executes the Verifier's verification protocol.
For example, when re-executing a training segment to verify that it would reach a particular chip-logged weight shard snapshot $\omega_{i+k}$, the cluster starts at one weight snapshot $W_i$, and then computes optimizer updates $k$ times using the specified inputs (where each update is computed via code generated from the hyperparameters, and even code snippets, defined by $\mathbb{M}$).
Finally, the cluster checks that the resulting weight vector $W'_{i+k}$ has a slice $\omega'_{i+k}$ that is within an $\epsilon$ distance of the chip-logged weight shard $\omega_{i+k}$.
If all the verification protocols passed, the cluster outputs that the Prover has ``passed''.
If not, the cluster outputs that the Prover ``failed'', prompting an investigation.

Assuming the training transcript is verified as correct, the Verifier can now compute any functions of the training transcript that would determine its compliance with agreed rules.
This could include properties of its training data distribution (which can be established from a randomly-selected subset of the training transcript's data), the performance of the final model on specific benchmarks, and properties of the hyperparameters.
Assuming the results confirm the Prover's compliance, the Verifier could be certain that with probability at least $1 - p - \delta_2$, the Prover has not used ML chips to execute a training runs using greater than $\rulecomputeflops$ FLOPs which violated the agreed upon rules.
The Prover has also not disclosed any sensitive information to the Verifier, including training data, model weights, or hyperparameters.

\end{document}